\documentclass[sigconf]{acmart}

\AtBeginDocument{%
  \providecommand\BibTeX{{%
    \normalfont B\kern-0.5em{\scshape i\kern-0.25em b}\kern-0.8em\TeX}}}

\setcopyright{acmcopyright}
\copyrightyear{2018}
\acmYear{2018}
\acmDOI{XXXXXXX.XXXXXXX}

\renewcommand\footnotetextcopyrightpermission[1]{}

\acmConference[Technical Report]{Make sure to enter the correct
  conference title from your rights confirmation emai}{ACM-MM Challenge}{2022}
\acmPrice{15.00}
\acmISBN{978-1-4503-XXXX-X/18/06}

\usepackage[linesnumbered,ruled,vlined]{algorithm2e}
\usepackage{mathrsfs}
\usepackage{booktabs}
\usepackage{multirow}
\usepackage{graphicx}  %插入图片的宏包
\usepackage{float}  %设置图片浮动位置的宏包
\usepackage{subfigure}  %插入多图时用子图显示的宏包

\begin{document}

\title{Team PKU-WICT-MIPL PIC Makeup Temporal Video Grounding Challenge 2022 Technical Report}

\author{Minghang Zheng \quad Dejie Yang \quad Zhongjie Ye \quad Ting Lei \quad Yuxin Peng \quad Yang Liu }
\authornote{Corresponding author}
\affiliation{%
  \institution{Wangxuan Institute of Computer Technology, Peking University}
  \city{Beijing}
  \country{China}
}
\email{{minghang, yangliu}@pku.edu.cn}

% \affiliation{%

\renewcommand{\shortauthors}{Minghang Zheng et al.}

\begin{abstract}

In this technical report, we briefly introduce the solutions of our team `PKU-WICT-MIPL' for the PIC Makeup Temporal Video Grounding (MTVG) Challenge in ACM-MM 2022. Given an untrimmed makeup video and a step query, the MTVG aims to localize a temporal moment of the target makeup step in the video. To tackle this task, we propose a phrase relationship mining framework to exploit the temporal localization relationship relevant to the fine-grained phrase and the whole sentence. Besides, we propose to constrain the localization results of different step sentence queries to not overlap with each other through a dynamic programming algorithm. The experimental results demonstrate the effectiveness of our method. Our final submission ranked 2nd on the leaderboard, with only a 0.55\% gap from the first.
 
 %In this report, we present a solution to the Makeup Temporal Video Grounding (MTVG) Challenge 2022.
\end{abstract}

\begin{CCSXML}
    <ccs2012>
        <concept>
        <concept_id>10010147.10010178.10010224.10010225.10010231</concept_id>
        <concept_desc>Computing methodologies~Visual content-based indexing and retrieval</concept_desc>
        <concept_significance>500</concept_significance>
    </concept>
    <concept>
        <concept_id>10010147.10010178.10010224.10010225.10010228</concept_id>
        <concept_desc>Computing methodologies~Activity recognition and understanding</concept_desc>
        <concept_significance>300</concept_significance>
        </concept>
     </ccs2012>
\end{CCSXML}

\ccsdesc[500]{Computing methodologies~Visual content-based indexing and retrieval}
\ccsdesc[300]{Computing methodologies~Activity recognition and understanding}

\keywords{Temporal Sentence Grounding,
Natural language query,
Sentence Localisation}

\maketitle

\section{Introduction}

Given an untrimmed make-up video and a step query, the Makeup Temporal Video Grounding (MTVG)~\cite{wang2019youmakeup} task aims to localize a temporal moment of the target make-up step in the video. This task requires models to align fine-grained video-text semantics and distinguish make-up steps with a subtle difference. YouMakeUp~\cite{wang2019youmakeup} dataset has two characteristics: (1) The step query is mainly composed of the actions (e.g., apply powder), the tools (e.g., with the brush), and the face regions (e.g., on the eyelids), which requires the model to understand the videos and queries in a more fine-grained way. (2) Make-up steps are done step by step, which means that the temporal moments corresponding to different steps do not overlap. However, most of the existing temporal video grounding methods only localize the entire query in a coarse-grained way. They cannot constrain the localization results of different steps not to overlap.

To alleviate these problems, we propose to mine the phrase-level relationship in the query in a fine-grained way, and constrain the localization results of different steps to not overlap through a dynamic programming algorithm. Specifically, we extract the phrases in the step query (e.g. `apply powder,' `on the eyelids,' and `with the brush'), simultaneously localize the step query and phrases, and comprehensively combine their localization results. In addition, during inference, we design a dynamic programming algorithm to select optimal temporal proposals for all step queries in a video and ensure that they do not overlap.

\section{Methodology}
% \begin{figure*}
%     \centering
%     \includegraphics[width=\linewidth]{figures/pipeline.pdf}
%     \caption{The overall framework of our approach.}
%     \label{fig:pipeline}
% \end{figure*}

We first extract the video features and build a 2D temporal feature map following MMN~\cite{DBLP:journals/corr/abs-2109-04872}. Then, we extract phrases in the step query, encode them separately, and interact between the sentence and phrases through a transformer encoder. Finally, each sentence and phrase feature will predict an importance weight and calculate the cosine similarity with the 2d temporal feature map to obtain the sentence and phrase score maps. The importance weights are used to fuse the phrase and sentence score maps and output the final score map. We train our network with binary cross-entropy loss and cross-modal mutual matching loss following MMN~\cite{DBLP:journals/corr/abs-2109-04872}, and further introduce an exclusiveness loss to encourage different step query predictions to be different from each other. In addition, during inference, we design a dynamic programming algorithm to select optimal temporal proposals for all step queries in a video and ensure that they do not overlap.

\subsection{2D Temporal Feature Map Encoder}

We use pretrained CLIP~\cite{radford2021learning} model (ViT-L/14@336px) to extract the frame features (roughly three frames of CLIP features per second). We also trim the videos by step queries and fine-tune the CLIP4Clip~\cite{luo2021clip4clip} model with the video retrieval task. We fine-tune the CLIP4Clip model initialized by the pretrained CLIP ViT-B/32 weight with learning rate equals to 0.0001 and batch size equals to 64 for 5 epochs. We found that many makeup actions can be inferred by comparing the changes of the face regions before and after makeup. Thus, to help the model better understand the structure of human face,
% \yang{you need to explain why model needs to understand face structure}
we introduce FaRL~\cite{zheng2021farl} face region features. For each frame of the CLIP or CLIP4Clip features, we simply average pool the face region features and concatenate them along channel dimension as the final video inputs. Finally, we build the 2D temporal feature map $V\in \mathbb{R}^{N\times N \times D}$ following MMN~\cite{DBLP:journals/corr/abs-2109-04872}, where $N=128$ is the number of video clips, $D=256$ is the feature dimension.

\subsection{Phrase Extraction and Query Encoder}

The step queries on YouMakeUp dataset are mainly composed of the actions (eg. apply powder), the tools (eg. with brush), and the face regions (eg. on the eyelids). In order to understand makeup step queries in a more fine-grained way, we use off-the-shelf SRLBERT~\cite{shi2019simple} model to extract phrases from step descriptions. SRLBERT assigns semantic rule labels to each word in the sentence, which serve as our phrases. Each phrase and the step sentence query will be encoded by CLIP~\cite{radford2021learning} text encoder separately. Finally, we use a 3-layer transformer encoder~\cite{vaswani2017attention} to interact between the sentence query and phrases. The final sentence feature is denoted as $T^s \in \mathbb{R}^{D}$, and the phrase feature is denoted as $T^p\in \mathbb{R}^{N_p\times D}$, where $N_p$ is the number of phrases.

\subsection{Phrase Similarity Learning}
Instead of only localizing the step sentence query, we also localize the phrases in a fine-grained way. We compute the cosine similarity of the sentence/phrase features and the 2d temporal feature map respectively and obtain the score maps, which represent the matching score between the temporal proposals and the sentence/phrase:
\begin{equation}
    S^s = \frac{VT^{s\top}}{\|V\|\|T^s\|}
\end{equation}
\begin{equation}
    S^p_i = \frac{VT_i^{p\top}}{\|V\|\|T_i^p\|}, i=1,2,...,N_p
\end{equation}
where $S^s\in \mathbb{R}^{N\times N}$ is the sentence score map, and $S^p\in \mathbb{R}^{N_p\times N\times N}$ is the phrase score maps. To measure the importance of sentence and phrases, we use a fully connected network to predict importance weights $w^s\in \mathbb{R}, w^p\in \mathbb{R}^{N_p}$ with text features $T^s, T^p$. We use a softmax activation function to ensure that these weights sum to 1. Finally, the importance weight $w^s, w^p$ will be used to perform a weighted summation of the score maps $S^s, S^p$ and output the fused score map $S$:
\begin{equation}
    S = w^sS^s+\sum_{i=1}^{N_p} w^p_iS^p_i \in \mathbb{R}^{N\times N}
\end{equation}

\subsection{Loss}
We train our network with binary cross entropy loss $\mathcal{L}_{bce}$ and cross-modal mutual matching loss $\mathcal{L}_{mm}$ following MMN~\cite{DBLP:journals/corr/abs-2109-04872}. The binary cross entropy loss uses the interaction over union (IoU) between the ground-truth and the proposals to supervise the score map $S$, and the cross-modal mutual matching loss contrast the positive moment-sentence pairs with the negative ones sampled from
both intra and inter videos. We further introduce the exclusiveness loss $\mathcal{L}_{exc}$ to encourage different step query predictions to be different from each other. The exclusiveness loss find the top-2 matching scores for each proposal from all the step queries in a video, and require their product to be 0. The exclusivity loss encourages any proposal to have a high matching score with at most one step query. The total loss of our method is shown blow:
\begin{equation}
    \mathcal{L} = \mathcal{L}_{bce} + \alpha \mathcal{L}_{mm} + \beta \mathcal{L}_{exc}
\end{equation}

\subsection{Inference with Dynamic Programming}

\begin{algorithm}[t] \SetKwInOut{Input}{input}\SetKwInOut{Output}{output}
	\Input{The number of steps $K$, the number of video clips $N$, and the log probability scores ${S_1,S_2,..,S_K}\in \mathbb{N\times N}$} 
% 	\Output{A partition of the bitmap}
	 \BlankLine 
	 $f_{\mathbb{K}, n}\leftarrow-\infty, \mathbb{K} \in \mathcal{P}(\{1,2,...,K\}), n\in\{1,2,...N\}$\;
	 \tcp{$f_{\mathbb{K}, n}$ represents the maximum score when allocate nonoverlapping proposals whose end time belongs to the $n$-th clip to the queries in the set $\mathbb{K}$}
	 $g_{\mathbb{K}, n}\leftarrow-\infty, \mathbb{K} \in \mathcal{P}(\{1,2,...,K\}), n\in\{1,2,...N\}$\;
	 \tcp{$g$ is the maximum value in the prefix of $f$ i.e. $g_{\mathbb{K}, n}=\max_{s\in{1,2,...,n}}f_{\mathbb{K}, s}$}
	 \For{$n\leftarrow 1$ \KwTo $N$}{
	    \For{$\mathbb{K} \in \mathcal{P}(\{1,2,...,K\})$}{
	       $f_{\mathbb{K}, n} \leftarrow \max_{k\in\mathbb{K}, s\in\{1,2,...,n\}} \left(g_{\mathbb{K}-\{k\}, s-1}+S_{k,s,n}\right)$\;
	    }
	    \For{$\mathbb{K} \in \mathcal{P}(\{1,2,...,K\})$}{
	       $g_{\mathbb{K}, n} \leftarrow \max\left(g_{\mathbb{K}, n-1}, f_{\mathbb{K}, n}\right)$\;
	    }
	}
	$\mathbb{K} \leftarrow \{1,2,...,K\}$\;
    $n \leftarrow \arg \max_{n\in\{1,2,...,N\}}  f_{K, n}$\;
    \While{$\mathbb{K}\ne \emptyset$}{
        $k,s\leftarrow \arg_{k\in \mathbb{K}, s\in\{1,2,...,n\}} \text{s.t.} f_{\mathbb{K}, n} = g_{\mathbb{K}-\{k\}, s-1}+S_{k,s,n} $\;
        \Output{The $k$-th step correspond to the video clips [s,n]}
        $\mathbb{K} \leftarrow \mathbb{K}-\{k\}$\;
        $n \leftarrow \arg \max_{n\in\{1,2,...,s-1\}}  f_{K, n}$\;
    }
    \caption{Inference with dynamic programming}
    \label{alg:dp} 
\end{algorithm}

As makeup steps are done step by step, the temporal moments corresponding to different steps do not overlap. During inference, we design a dynamic programming algorithm to select optimal temporal proposals for all step queries in a video and ensure that they do not overlap. Specifically, we regard the predicted scores for each proposal as the probability of choosing that proposal. Let $P(k, s, n)$ denote the probability of choosing the proposal $(s, n), 1\le s\le n\le N$ for the $k$-th step query and $K$ denote the total number of steps. Our goal is to find a sequence of nonoverlapping proposals $(s_1,n_1),(s_2,n_2),...,(s_K,n_K)$ that maximizes the joint probability when the proposal $(s_i,n_i)$ is chosen for the $i$-th step query. That is:
\begin{equation}
    \{(s_i, n_i)\} = \arg\max_{\{(s_i,n_i)\}}\Pi_iP(i, s_i, n_i)
\end{equation}
satisfy:
\begin{equation}
    \exists \sigma \in \mathcal{P}_K \text{ s.t. } s_{\sigma_1}\le n_{\sigma_1}< s_{\sigma_2}\le n_{\sigma_2}<...< s_{\sigma_K}\le n_{\sigma_K}
\end{equation}
where $\mathcal{P}_K$ represents the set of the permutations of $K$ elements. We use the logarithmic function to turn multiplications into additions in the joint probability calculation and use the dynamic programming algorithm in Algorithm~\ref{alg:dp} to solve the above optimization problem. Since the complexity of the algorithm is $O(2^KN^2K)$, we will only execute the algorithm for those samples with $K \le 17$

\section{Experiments}

\begin{table}[t]
    \centering
    \begin{tabular}{c|c|cccc}
        \toprule
         Features & Methods & IoU=0.3 & IoU=0.5 & IoU=0.7 & AVG\\
         \midrule
         \multirow{3}{*}{CLIP} & MMN~\cite{DBLP:journals/corr/abs-2109-04872} & 59.43 & 46.73 & 27.57 & 44.58 \\
         & Ours-base & 60.15 & 47.46 & 28.26 &45.29 \\
         & +DP & \textbf{62.07} & \textbf{49.79} & \textbf{29.68} & \textbf{47.18} \\
         \midrule
         \multirow{3}{*}{CLIP4Clip} & Our-base & 68.42 & 53.90 & 34.07 & 52.13 \\
         & Ours-exc & 68.42 & 54.31 & 33.85 & 52.19 \\
         & Ours-exc-f & 69.59 & 53.74 & 33.12 & 52.15 \\
         & +DP & \textbf{70.22} & \textbf{56.83} & \textbf{34.73} & \textbf{53.93}  \\
        \midrule
         & Ensemble & 72.12 & 59.77 & 40.29 & 57.39 \\
         & +DP & \textbf{75.56} & \textbf{63.21} & \textbf{42.25} & \textbf{60.34} \\
        \bottomrule
    \end{tabular}
    \caption{Performance of different methods on the val set.}
    \label{tab:val}
\end{table}

\begin{table}[t]
    \centering
    \begin{tabular}{c|cccc}
        \toprule
         Methods & IoU=0.3 & IoU=0.5 & IoU=0.7 & AVG\\
         \midrule
         Ensemble+DP & 73.61 & 62.50 & 42.12 & 59.41 \\
        \bottomrule
    \end{tabular}
    \caption{Performance of our ensemble model on test set.}
    \label{tab:test}
\end{table}

As shown in Tab.~\ref{tab:val}, we compare our model with baseline MMN~\cite{DBLP:journals/corr/abs-2109-04872}. We denote our model without exclussiveness loss as Ours-base, denote our model with exclussiveness loss as Ours-exc, and denote the full model with face region feature as Ours-exc-f. `+DP' represents inference with our dynamic programming algorithm. `Ensemble' represents ensemble the all the models (without DP) by averaging the predicted score map. 

Tab.~\ref{tab:val} shows the results on the validation set. As we can see, (1) fine-grained consideration of phrases and introduction of exclussiveness loss can improve the performance. (2) During inference, the dynamic programming algorithm ensures that the predictions do not overlap, and can improve performance by about 2\%. (3) The finetuned CLIP4Clip feature has a better performance than the CLIP feature, which can improve the performance by about 6.8\%. (4) Through model ensemble, the performance can be improved by 6.4\%. The test set performance of our ensemble model with dynamic programming achieves the best performance on IoU=0.3 and IoU=0.5 and the second performance on IoU=0.7 and AVG, as shown in Tab.~\ref{tab:test}.

\section{Qualitative Results}

\begin{figure*}[t]
    \centering
    \subfigure[Successful Case]{
		\label{fig:vis_1}
		\includegraphics[width=0.9\linewidth]{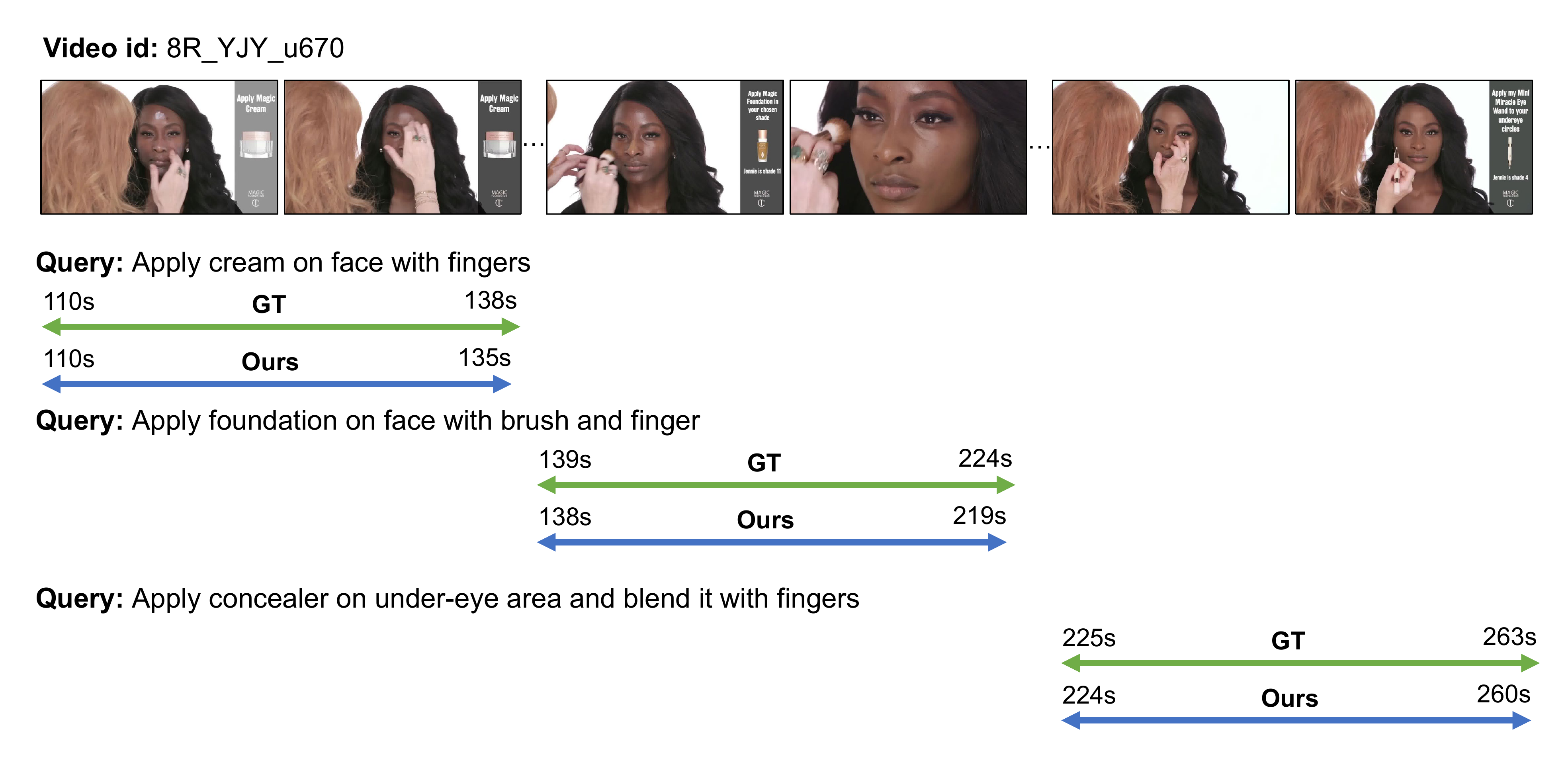}
	}
    \subfigure[Failure Case]{
		\label{fig:vis_2}
		\includegraphics[width=0.9\linewidth]{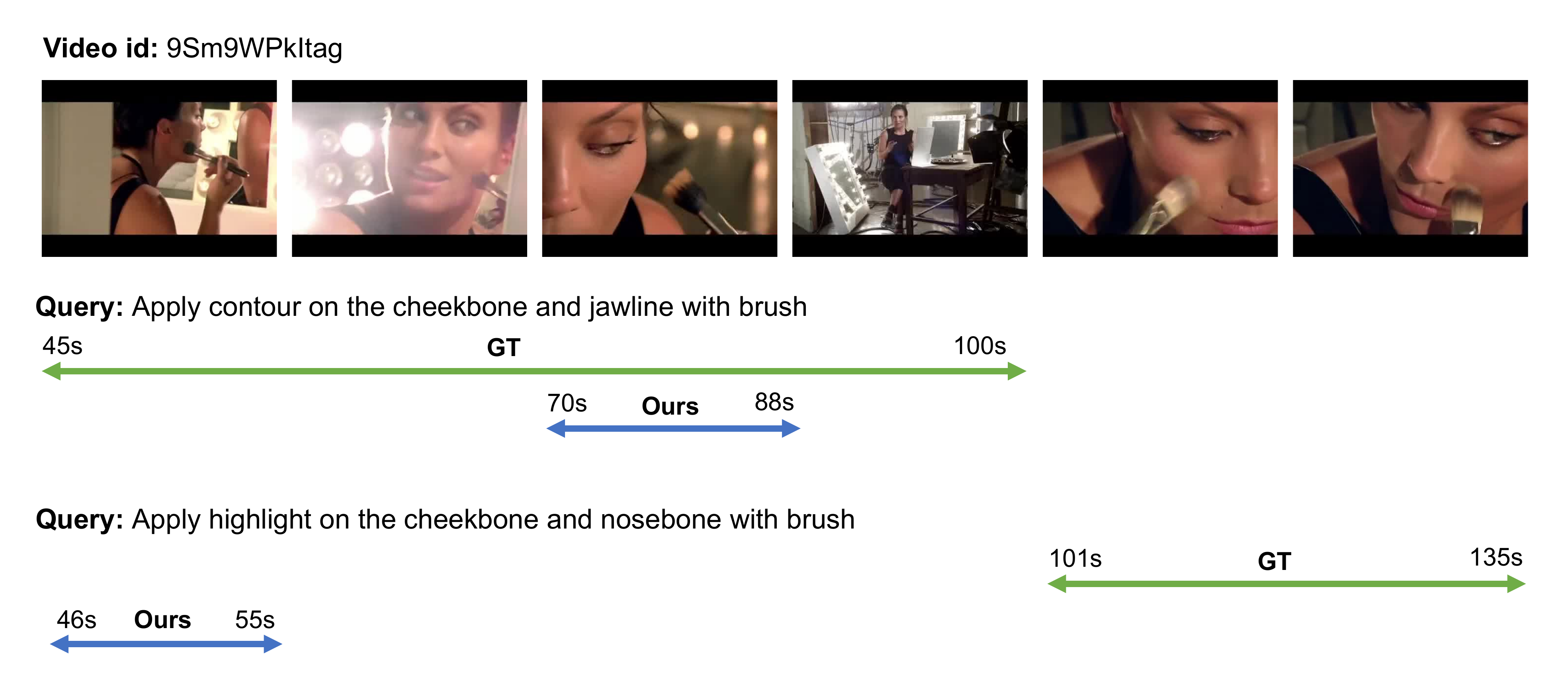}
	}
    \caption{Qualitative results on the validation set.}
    \label{fig:vis}
\end{figure*}

Fig.~\ref{fig:vis} shows some qualitative results of our model on the validation set. 
% As we can see, our dynamic programming algorithm can ensure the predictions do not overlap. 
% \yang{We can see they do not overlap, but cannot know whether the reason behind is using DP. You can either(1) draw prediction before and after usinig DP (2) report the overlap stats before and after to support your argument here} 
According to our statistics, with dynamic programming only performed on videos containing no more than 17 queries, we reduced the proportion of queries that overlapped with other predictions from 49.1\% to 17.7\%, demonstrating the effectiveness of the dynamic programming algorithm in minimizing the overlap.
In Fig.~\ref{fig:vis_1}, our model successfully predicted all queries correctly. In Fig.~\ref{fig:vis_2}, the video clip corresponding to the first query is split into multiple shots, which results in our model predicting only one of the shots. At the same time, the two queries refer to the same tool (i.e., with the brush) and face region (i.e., cheekbone). The only difference is the makeup action (i.e., apply contour and apply highlight), yet the two actions are visually similar. Our model fails to understand the difference between these two actions at a fine-grained level and predicts the wrong answer.

\section{Conclusion}
In our submission to the PIC Makeup Temporal Video Grounding Challenge 2022, we explore the relationship  between the phrases in the query in a fine-grained way, and constrain the localization results of different steps to not overlap through a dynamic programming algorithm. Experiments show the effectiveness of our method and we win the 2nd place on the leaderboard, with only a 0.55\% gap from the first.

\begin{acks}
This work is supported by the grants from the National Natural Science Foundation of China (61925201, 62132001, U21B2025), Zhejiang Lab (NO. 2022NB0AB05) and Beijing Institute for General Artificial Intelligence (NO.A010121004).  
\end{acks}

\bibliographystyle{ACM-Reference-Format}
\bibliography{acmart}


\begin{thebibliography}{7}

%%% the Web address.

\ifx \showCODEN    \undefined \def \showCODEN     #1{\unskip}     \fi
\ifx \showDOI      \undefined \def \showDOI       #1{#1}\fi
\ifx \showISBNx    \undefined \def \showISBNx     #1{\unskip}     \fi
\ifx \showISBNxiii \undefined \def \showISBNxiii  #1{\unskip}     \fi
\ifx \showISSN     \undefined \def \showISSN      #1{\unskip}     \fi
\ifx \showLCCN     \undefined \def \showLCCN      #1{\unskip}     \fi
\ifx \shownote     \undefined \def \shownote      #1{#1}          \fi
\ifx \showarticletitle \undefined \def \showarticletitle #1{#1}   \fi
\ifx \showURL      \undefined \def \showURL       {\relax}        \fi
% invisible to TeX
\providecommand\bibfield[2]{#2}
\providecommand\bibinfo[2]{#2}
\providecommand\natexlab[1]{#1}
\providecommand\showeprint[2][]{arXiv:#2}

\bibitem[Luo et~al\mbox{.}(2021)]%
        {luo2021clip4clip}
\bibfield{author}{\bibinfo{person}{Huaishao Luo}, \bibinfo{person}{Lei Ji},
  \bibinfo{person}{Ming Zhong}, \bibinfo{person}{Yang Chen},
  \bibinfo{person}{Wen Lei}, \bibinfo{person}{Nan Duan}, {and}
  \bibinfo{person}{Tianrui Li}.} \bibinfo{year}{2021}\natexlab{}.
\newblock \showarticletitle{Clip4clip: An empirical study of clip for end to
  end video clip retrieval}.
\newblock \bibinfo{journal}{\emph{arXiv preprint arXiv:2104.08860}}
  (\bibinfo{year}{2021}).
\newblock


\bibitem[Radford et~al\mbox{.}(2021)]%
        {radford2021learning}
\bibfield{author}{\bibinfo{person}{Alec Radford}, \bibinfo{person}{Jong~Wook
  Kim}, \bibinfo{person}{Chris Hallacy}, \bibinfo{person}{Aditya Ramesh},
  \bibinfo{person}{Gabriel Goh}, \bibinfo{person}{Sandhini Agarwal},
  \bibinfo{person}{Girish Sastry}, \bibinfo{person}{Amanda Askell},
  \bibinfo{person}{Pamela Mishkin}, \bibinfo{person}{Jack Clark},
  {et~al\mbox{.}}} \bibinfo{year}{2021}\natexlab{}.
\newblock \showarticletitle{Learning transferable visual models from natural
  language supervision}. In \bibinfo{booktitle}{\emph{International Conference
  on Machine Learning}}. PMLR, \bibinfo{pages}{8748--8763}.
\newblock


\bibitem[Shi and Lin(2019)]%
        {shi2019simple}
\bibfield{author}{\bibinfo{person}{Peng Shi} {and} \bibinfo{person}{Jimmy
  Lin}.} \bibinfo{year}{2019}\natexlab{}.
\newblock \showarticletitle{Simple bert models for relation extraction and
  semantic role labeling}.
\newblock \bibinfo{journal}{\emph{arXiv preprint arXiv:1904.05255}}
  (\bibinfo{year}{2019}).
\newblock


\bibitem[Vaswani et~al\mbox{.}(2017)]%
        {vaswani2017attention}
\bibfield{author}{\bibinfo{person}{Ashish Vaswani}, \bibinfo{person}{Noam
  Shazeer}, \bibinfo{person}{Niki Parmar}, \bibinfo{person}{Jakob Uszkoreit},
  \bibinfo{person}{Llion Jones}, \bibinfo{person}{Aidan~N Gomez},
  \bibinfo{person}{{\L}ukasz Kaiser}, {and} \bibinfo{person}{Illia
  Polosukhin}.} \bibinfo{year}{2017}\natexlab{}.
\newblock \showarticletitle{Attention is all you need}.
\newblock \bibinfo{journal}{\emph{Advances in neural information processing
  systems}}  \bibinfo{volume}{30} (\bibinfo{year}{2017}).
\newblock


\bibitem[Wang et~al\mbox{.}(2019)]%
        {wang2019youmakeup}
\bibfield{author}{\bibinfo{person}{Weiying Wang}, \bibinfo{person}{Yongcheng
  Wang}, \bibinfo{person}{Shizhe Chen}, {and} \bibinfo{person}{Qin Jin}.}
  \bibinfo{year}{2019}\natexlab{}.
\newblock \showarticletitle{YouMakeup: A Large-Scale Domain-Specific Multimodal
  Dataset for Fine-Grained Semantic Comprehension}. In
  \bibinfo{booktitle}{\emph{Proceedings of the 2019 Conference on Empirical
  Methods in Natural Language Processing and the 9th International Joint
  Conference on Natural Language Processing (EMNLP-IJCNLP)}}.
  \bibinfo{pages}{5136--5146}.
\newblock


\bibitem[Wang et~al\mbox{.}(2021)]%
        {DBLP:journals/corr/abs-2109-04872}
\bibfield{author}{\bibinfo{person}{Zhenzhi Wang}, \bibinfo{person}{Limin Wang},
  \bibinfo{person}{Tao Wu}, \bibinfo{person}{Tianhao Li}, {and}
  \bibinfo{person}{Gangshan Wu}.} \bibinfo{year}{2021}\natexlab{}.
\newblock \showarticletitle{Negative Sample Matters: {A} Renaissance of Metric
  Learning for Temporal Grounding}.
\newblock \bibinfo{journal}{\emph{CoRR}}  \bibinfo{volume}{abs/2109.04872}
  (\bibinfo{year}{2021}).
\newblock


\bibitem[Zheng et~al\mbox{.}(2021)]%
        {zheng2021farl}
\bibfield{author}{\bibinfo{person}{Yinglin Zheng}, \bibinfo{person}{Hao Yang},
  \bibinfo{person}{Ting Zhang}, \bibinfo{person}{Jianmin Bao},
  \bibinfo{person}{Dongdong Chen}, \bibinfo{person}{Yangyu Huang},
  \bibinfo{person}{Lu Yuan}, \bibinfo{person}{Dong Chen}, \bibinfo{person}{Ming
  Zeng}, {and} \bibinfo{person}{Fang Wen}.} \bibinfo{year}{2021}\natexlab{}.
\newblock \showarticletitle{General Facial Representation Learning in a
  Visual-Linguistic Manner}.
\newblock \bibinfo{journal}{\emph{arXiv preprint arXiv:2112.03109}}
  (\bibinfo{year}{2021}).
\newblock


\end{thebibliography}

\end{document}